\documentclass[letterpaper, 10 pt, conference]{ieeeconf}  

\IEEEoverridecommandlockouts                              

\usepackage{cite}
\usepackage{amsmath,amssymb,amsfonts}
\usepackage{graphicx}
\usepackage{textcomp}
\usepackage{xcolor}
\usepackage{bm}
\usepackage{mathtools}
\usepackage{float}
\let\labelindent\relax
\usepackage{enumitem}
\usepackage{afterpage}
\usepackage{array,booktabs,ragged2e}
\usepackage{longtable}
\usepackage{tabularx}
\usepackage{siunitx}
\usepackage{caption}
\usepackage{hyperref}
\usepackage{pifont}
\usepackage{booktabs}
\usepackage{algorithm}
\usepackage[noend]{algpseudocode}

\DeclareMathOperator*{\argmax}{arg\!\max}

\def\tick{\color{green}\ding{51}}
\def\cross{\color{red}\ding{55}}

\title{\LARGE \bf
Multi-fingered Dynamic Grasping for Unknown Objects
}

\author{Yannick Burkhardt$^{1,2,\dag}$, Qian Feng$^{1,2,\dag}$,
Jianxiang Feng$^{1,2}$, 
Karan Sharma$^{1}$, \\
Zhaopeng Chen$^{1}$, Alois Knoll$^{2}$
\thanks{$^{1}$Agile Robots AG, \href{mailto:qian.feng@agile-robots.com}{\nolinkurl{qian.feng@agile-robots.com}}}
\thanks{$^2$Department of Informatics, Technical University of Munich (TUM)}
\thanks{$^\dag$These authors contributed equally to this work.}%
}

\begin{document}

\maketitle
\thispagestyle{empty}
\pagestyle{empty}

\begin{abstract}
Dexterous grasping of unseen objects in dynamic environments is an essential prerequisite for the advanced manipulation of autonomous robots.
Prior advances rely on several assumptions that simplify the setup, including environment stationarity, pre-defined objects, and low-dimensional end-effectors.
Though easing the problem and enabling progress, it undermined the complexity of the real world. 
Aiming to relax these assumptions, we present a dynamic grasping framework for unknown objects in this work, which uses a five-fingered hand with visual servo control and can compensate for external disturbances. 
To establish such a system on real hardware, we leverage the recent advances in real-time dexterous generative grasp synthesis and introduce several techniques to secure the robustness and performance of the overall system.
Our experiments on real hardware verify the ability of the proposed system to reliably grasp unknown dynamic objects in two realistic scenarios: objects on a conveyor belt and human-robot handover. Note that there has been no prior work that can achieve dynamic multi-fingered grasping for unknown objects like ours up to the time of writing this paper. 
We hope our pioneering work in this direction can provide inspiration to the community and pave the way for further algorithmic and engineering advances on this challenging task. A video of the experiments is available at \href{https://youtu.be/b87zGNoKELg}{https://youtu.be/b87zGNoKELg}.
\end{abstract}

\afterpage{
\begin{figure}[t]
\includegraphics*[width=\columnwidth]{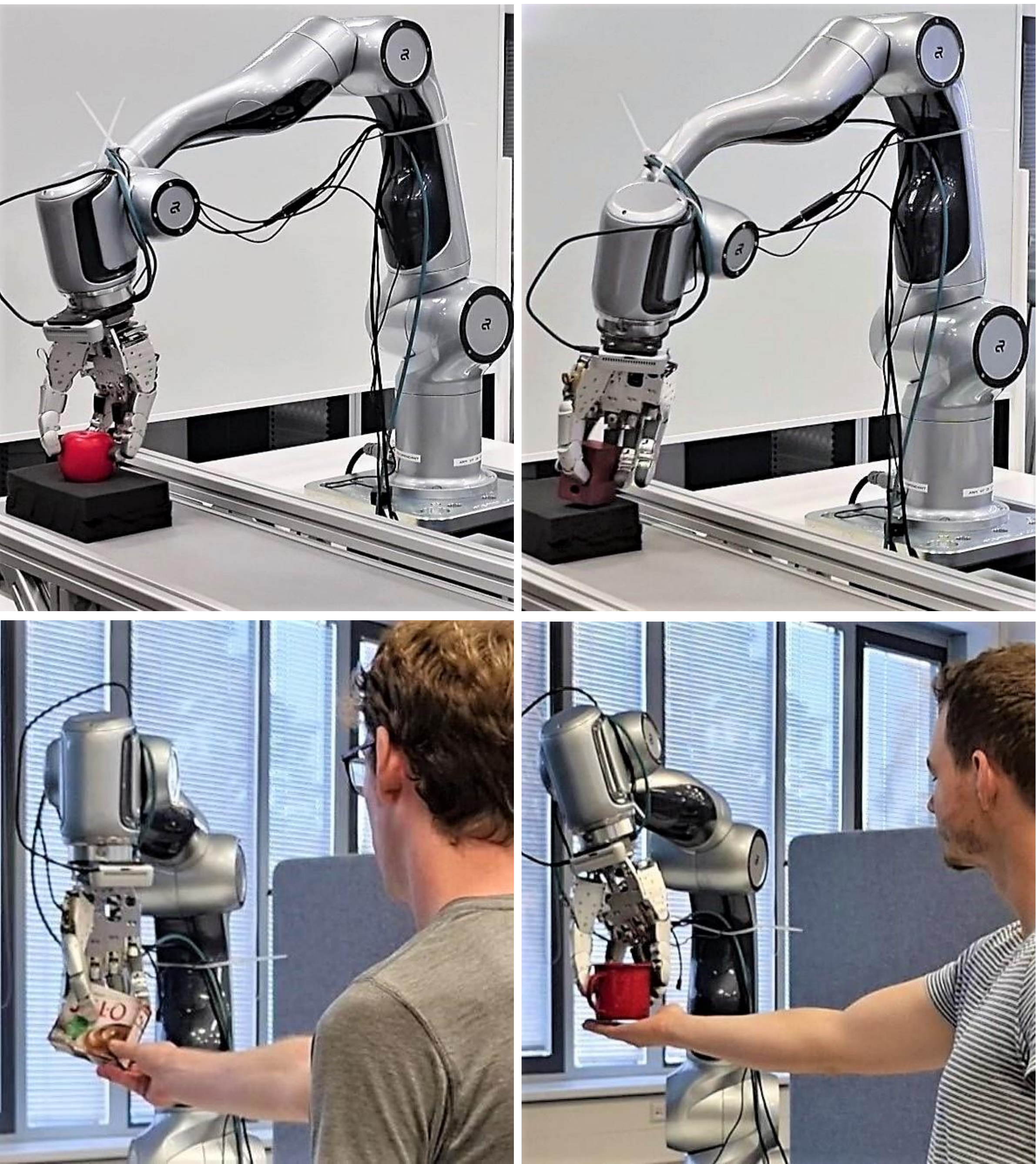}
\centering
\caption{\textbf{Multi-fingered dynamic grasping in two exemplar scenarios}. Successful grasps in the conveyor belt experiment with target velocities of \SI{220}{mm \per s} for the apple (top left) and \SI{180}{mm \per s} for the foam brick (top right). Human-to-robot handovers of the pudding box (bottom left) and the mug (bottom right).}
\label{exp_photos}
\end{figure}
}

\section{Introduction}
Dexterous robotic grasping of dynamic and unknown objects acts as a crucial precursor to advanced manipulation required by various tasks. 
It can be extremely important to realize automation in both industry and daily life. 
For example, Figure \ref{exp_photos} illustrates two potential scenarios: grasping objects on a conveyor belt and human-robot handover. 
Establishing such automated robotic solutions can significantly improve overall efficiency and productivity.

Towards this goal, prior works have made tremendous progress along with various assumptions that simplify the problem setting, including environment stationarity, predefined objects to interact with, and low-dimensional end-effectors such as the parallel-jaw gripper. 
Though these assumptions can ease the analysis and promote the research in this line, they meanwhile restrict the applicability of robotic grasping in the real world, which is often dynamic, full of unseen objects, and sometimes in need of complex manipulation. 
In light of this, we aim to relax these assumptions and tackle the problem of multi-fingered dynamic grasping that can handle unknown objects.

The challenges to achieving reliable robotic grasping of moving objects with a multi-fingered hand are three-fold based on our experience.
First, in contrast to two- or three-jaw grippers, a multi-fingered hand has much higher degrees of freedom (DoF), such as the 15 DoF anthropomorphic DLR-HIT Hand II~\cite{Liu08} used in this work, which presents \textit{greater complexity} in control.
Second, the grasp synthesis component for such high dimensional end-effectors that is repetitively executed according to the updated object pose needs to be \textit{real-time capable}.  
Third, different failure factors (e.g., tracking loss, inaccurate segmentation) during the grasping process need to be handled properly by \textit{considering the approach and close motion of the grasp}.

Previous works addressed these issues in dynamic grasping by introducing a multitude of assumptions ranging from the prior knowledge of object motion to a pre-collected database of objects with known geometry and hence, their grasps. 
In this work, we propose a dynamic grasping framework that can work with \textit{unseen objects} with \textit{no prior knowledge of object motion}.
Specifically, the proposed system consists of two major processes, namely, the target model generation and the grasp control. 
The first one processes the RGB-D data and maintains an internal model point cloud whose pose is updated at each time step.  
At the same time, depending on the internal point cloud, the grasp control module adaptively generates suitable grasps. 
It controls the robot to approach the predicted grasp and execute the grasp based on a carefully designed strategy. 
To facilitate real-time and adaptive grasp synthesis, we leverage a recent learning-based generative grasp synthesis model~\cite{May21}.
In addition, to increase the robustness of the online grasp generation, we make use of observations at multiple time steps by fusing them into the internal model point cloud. 
Furthermore, to facilitate stable grasp control, we design a customized metric for high-quality grasp filtering. 
The metric is carefully designed by considering both the geometric and semantic information of the predicted grasp.
With all combined in one system, our ultimate grasping framework visualized in Figure \ref{loop} can also handle the failure case of control feedback interruption. 
For example, due to the camera's limited field of view and the minimum depth range, the visual data is unreliable shortly before grasp execution, making an estimation essential when grasping moving objects.

To verify the proposed framework, we design two experimental scenarios that closely simulate the tasks in the real world, namely grasping objects on a conveyor belt and human-robot handover.  
To thoroughly examine the robustness of grasping objects on the conveyor belt, we evaluate the system with various moving speeds of the objects.
In the human-robot handover experiment, we solicit ten participants uniformly. 
Though they intentionally help the robot grasp the object, the unpredictable motion patterns from different people pose large difficulties for the system. 
Nevertheless, the encouraging results from both experiments empirically prove the effectiveness of our framework.

\begin{table*}[t]
\begin{center}
\vspace{5pt}
\caption{Comparison between dynamic grasping approaches that generate or adjust grasps online.\label{related}}
\begin{tabular*}{\linewidth}{@{\extracolsep{\fill}} lcccccccccc }
\toprule
 & \cite{Mor18, Mor19} & \cite{Tus21} & \cite{Mar19} & \cite{deF21} & \cite{Aki21} & \cite{Won22} & \cite{Fan23} & \cite{Ros21} & \cite{Yan21} & \textbf{Ours} \\
\midrule
Multi-fingered grasping ($>$ 1 DoF) & \cross & \cross & \cross & \cross & \cross & \cross & \cross & \cross & \cross & \tick \\

Online grasp generation & \tick & \cross & \cross & \cross & \cross & \cross & \cross & \cross & \tick & \tick \\

Real-time control (loop frequency $\geq$ \SI{30}{Hz}) & \tick & \tick & \tick & \tick & \tick & \tick & \cross & \tick & \cross & \tick \\

Low initialization time for unseen objects ($<$ \SI{10}{s}) & \tick & \tick & \cross & \cross & \cross & \cross & \tick & \tick & \tick & \tick \\

Object movement during grasp execution & \cross & \cross & \tick & \tick & \tick & \tick & \tick & \cross & \cross & \tick \\

Full 6 DoF Grasping & \cross & \tick & \tick & \tick & \tick & \cross & \tick & \tick & \tick & \tick \\

\bottomrule
\vspace{-15pt}
\end{tabular*}
\end{center}
\end{table*}

To summarize, the contributions of this work are: 
\setlist{nolistsep}
\begin{itemize}[noitemsep, leftmargin=*]
\item To the best of our knowledge, we are the first to tackle this challenging problem of multi-fingered dynamic grasping for unknown objects and introduce a practical framework that can run efficiently on real hardware.

\item Within the proposed system, we devise effective techniques to achieve such a demanding task, including a hybrid metric to filter grasp and a specific strategy for approach and close motion of the robot. 
This allows our system to robustly handle the failure case of control feedback interruption.
\item We perform empirical experiments on real hardware for two representative tasks, namely grasping objects on a conveyor belt and human-robot handover. 
The quantitative results demonstrate the effectiveness and robustness of our system. 
Additionally, we show qualitative results in a video and conduct a detailed failure analysis.
\end{itemize}

\section{Related Work}
The majority of prior works is centered around two or three-jaw grippers~\cite{Hus14, Ye18, Pha18, Tu22}. 
Additionally, they impose restrictions on the target object to be graspable with this finger configuration. 
Instead, we employ a multi-fingered hand with superior manipulation capabilities and enable the grasping of truly unknown objects.\par
\textbf{Pre-programmed grasp execution:} Early works are limited to executing pre-programmed grasps for a single known object. While initially, the target's trajectory must be known a priori~\cite{Hou90, All93}, this simplification is later omitted~\cite{Smi95, Nom00, Fue09}. Also, more recent approaches constrain the target and its pose to ensure the stability of a pre-programmed grasp~\cite{Cow13, Esc14, Hus14, Ye18}. Some additionally require the target to be known or to fulfill specific geometric properties~\cite{Cow13, Esc14, Ye18}. Another approach greatly simplifies visual tracking and grasp planning by marking the grasp pose with an Apriltag on the target~\cite{Aro20}.\par

\textbf{Online grasp generation or adjustment:}
Recent works use data-driven models to generate grasps. 
Some rely only on the most recent observation and discard all structural information from previous time steps~\cite{Mor18, Mor19, Tus21}. 
The quality of camera data cannot be guaranteed in case of occlusions, unclear perspectives of the target, or inaccurate sensor feedback. 
While these implementations can handle changes in the target pose during approaching, they require the target to be static directly before and during the grasp execution. 
In contrast, our online generation of the target model enables the system to improve its grasp predictions and to estimate the target's translational velocity in case of tracking loss. 
Additionally, the visual servoing framework in GGCNN~\cite{Mor18, Mor19} has only 4 DoF, while ours generates a full 6 DoF grasp. 
The grasp generation module in \cite{Tus21} is only designed for cuboid objects. In another approach, the object's model is not created online but requires a preceding time-inefficient scanning phase~\cite{Mar19}. 
Different from us, some systems~\cite{Mar19, deF21, Aki21} imply a priori knowledge of the target's geometrical properties and use pre-generated sets of grasps.
These grasps are continuously re-ranked during object tracking based on the target object's movement. 
Also, the pre-processing step to generate grasps significantly increases process times. 
Our framework can continuously complete the target model and generate grasps online without a scanning phase.
Moreover, our system does not rely on known or familiar objects or a grasp database.\par
Recent work to enable dynamic grasping by predicting the target's movement does not generalize to unknown objects nor arbitrary trajectories~\cite{Won22}. 
The system AnyGrasp cannot adjust the grasp online based on visual feedback, and its speed of $\SI{7}{Hz}$ violates the real-time constraint, restricting the target's movements to be slow and predictive~\cite{Fan23}. 
In contrast, our approach regenerates grasps during trajectory execution, enabling a more adaptive reaction to external disturbances. Additionally, the more complete virtual object representation improves grasp prediction.\par 
Table~\ref{related} compares the existing approaches that generate or update the grasp online to our implementation.\par

\textbf{Reinforcement learning (RL) based approaches:}
Other approaches apply RL to the dynamic grasping problem~\cite{Pha18, Che21, Tu22, Hua24}. However, they all greatly reduce the grasp generation complexity by only considering spherical, cubic, or other known objects. Additionally, some further simplify the problem by either employing Apriltags for object detection~\cite{Pha18}, reporting a significant gap in the grasp success rate between simulated and real experiments~\cite{Che21}, or exclusively evaluating their model in simulation~\cite{Tu22}.\par
Our model does not require time-consuming and inflexible pre-training for specific objects or any of the mentioned simplifications.

\section{Dynamic Grasping Framework}
\begin{figure*}[ht!]
    \centering

    \includegraphics[width=1.0\textwidth, height=9.1cm]{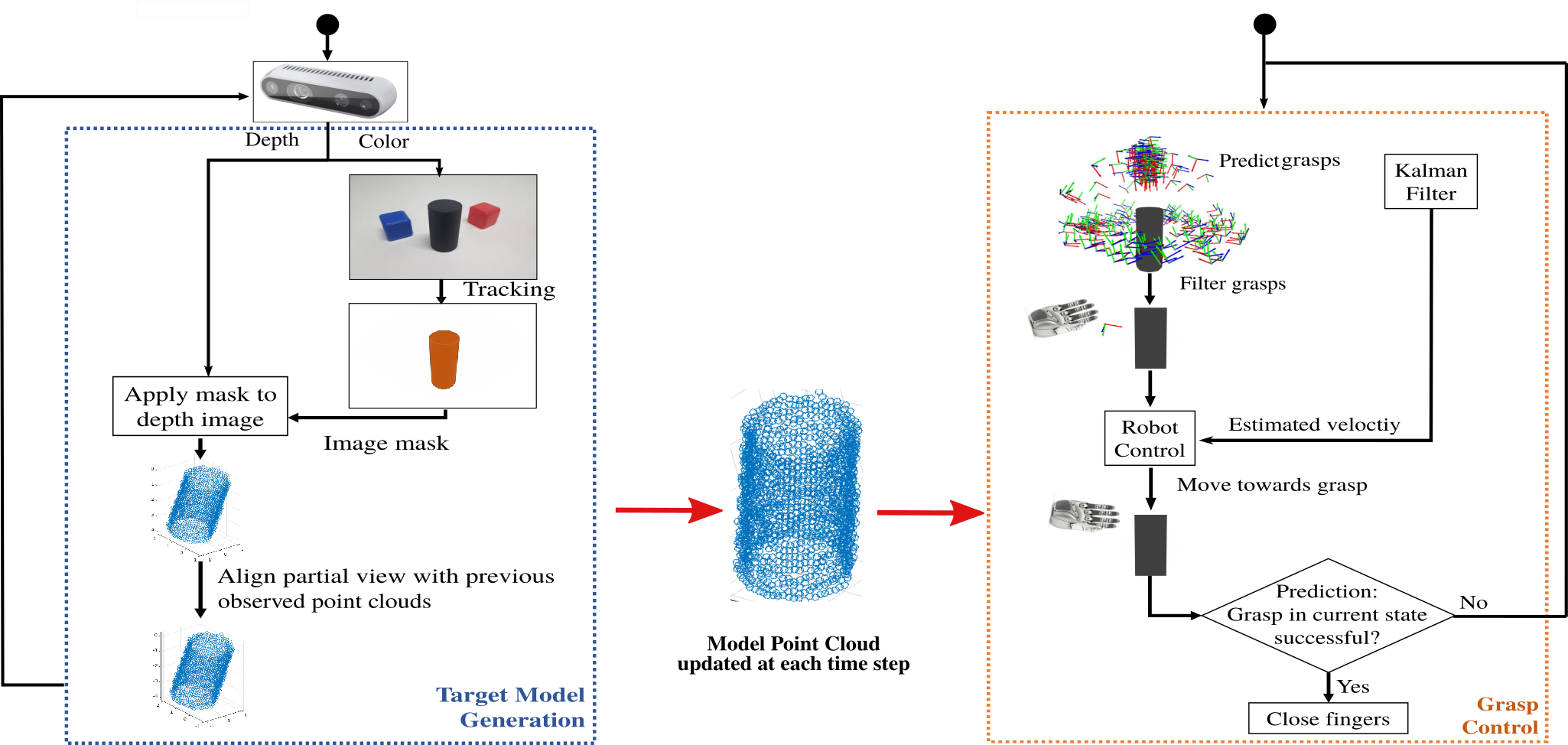}
    \caption{\textbf{Schematic representation of the proposed dynamic grasping framework.} Our system consists of two asynchronous processes, namely target model generation and grasp control. The former is responsible for maintaining an internal point cloud representation of the target object model based on the RGB-D data. The latter achieves adaptive grasping based on the latest internal model point cloud.
    } 
    \label{loop}
    \vspace{-2pt}
\end{figure*}

The control system is divided into the two processes \textit{Target Model Generation} and \textit{Grasp Control}.
As illustrated in Figure~\ref{loop}, the \textit{Target Model Generation} process fuses recent frames of partial views for the purpose of the whole object model generation.
Meanwhile, the \textit{Grasp Control} process determines a suitable grasp and moves the robot towards it until the current hand pose is predicted to allow successful grasping.
They run asynchronously to avoid blocking and to enable efficient recovery after errors such as tracking loss. 
Figure~\ref{loop} shows the procedural steps of the two processes, which are described in detail in the following.

\subsection{Target Model Generation}\label{tpcp}
\subsubsection{Tracking}
The camera feedback is processed to create a point cloud model representing the target object.
The depth camera attached to the robot's end-effector (eye-in-hand configuration) provides a color and depth data stream $\mathbf{O_t = (RGB, D)}$. 
A visual object tracker segments the target object from subsequent color images as long as it stays in the camera's observable space based on the transformer-based model \textit{TransT-M}~\cite{Che21trans}. 
As the winner of the VOT-RT2021~\cite{Kri21} challenge, TransT-M offers state-of-the-art robustness and accuracy for a vast diversity of objects while obeying real-time constraints. 
We manually initialize the tracker with a bounding box around the target object to keep the system independent of specific object classes, e.g., a semantic segmentation data set. 
Note that this bounding box can also be initialized by an open-world object detector, which does not limit the suitability of our approach. 

Upon alignment of the camera's color and depth data, the segmentation mask provided by the tracker is applied to the depth image, resulting in the segmentation of the target object's surface structure. 
This depiction is converted to a point cloud. 
In order to avoid confusion, this point cloud directly generated from the camera data is called \textit{observation} point cloud $\mathbf{P}_t$ at time step $t$. The point cloud created by post-processing and merging the observation point clouds is called \textit{model} point cloud $\mathbf{Q}_t$. The model point cloud represents the object from all observed perspectives.\par

\subsubsection{Point Cloud Post-processing}
First, the observation point cloud is post-processed by removing outliers that belong to the background and are far away from the center. 

Outliers are filtered out by removing all points whose $z$-coordinates $p_{t,i,z}$ differ from the median depth of the observation point cloud ${P}_t$ by more than a threshold $c_z$, in camera's $z$-axis.\par


The transformation to align the observation point clouds from different time steps without a priori knowledge of the target movement can only be inferred by employing their structural information. This is achieved by registration with the \textit{Iterative Closest Point} (ICP) algorithm~\cite{Bes92, Che92}, which allows a fast approximation of the transformation between two point clouds by iteratively maximizing their overlap.\par
However, applying a local optimization algorithm such as ICP to noisy input data is error-prone. A false estimation can lead to imprecise grasps since the resulting transformation is required to transform the model point cloud into the current camera frame. To distinguish successful from unsuccessful ICP alignment, two criteria must be fulfilled:\\
(1)\hspace{1mm}The alignment fitness score exceeds the threshold $c_{a,f}$.\\
(2)\hspace{1mm}The resulting translation and rotation norms between the two centered point clouds remain smaller than an upper bound. This implies the maximal relative linear velocity $c_{a,\hat{v}}$ and rotational velocity $c_{a,\hat{\omega}}$ between the camera and the target.\par

\begin{algorithm}
\caption{Dynamic Grasping Framework\label{alg:cap}}
\begin{algorithmic}[1]
\Require A color and depth frame $\mathbf{O}_t = (\mathbf{RGB, D})$ at time step $t$ whose first frame is initialized with a user-specified bounding box; A threshold $thresh$ for determining grasp execution. 
\textbf{Note:} We assume no tracking loss and no ICP failure for the first $\tau$ frames. The two following functions run asynchronously.

\Function {Target Model Generation} {$\mathbf{O}_t$}
\While{True\do} 
    \State $P_t \gets Tracking(\mathbf{O}_t)$
    \State ${P}''_t \gets PointCloudFilter(P_t)$
    \If{$t=0$}
        \State $Q_0 \gets {P}''_0$
    \Else
        \State ${Q}_t \gets MergePointCloud(Q_{t-1},{P}''_t)$
    \EndIf
    \State{Return ${Q}_t$}
\EndWhile
\EndFunction

\Function {Grasp Control} {${Q}_t$}
\While{True\do} 
    \If{$No TrackingLoss$ and $No ICP Failure$ \do}
        \State $\mathbb{G}_i \gets FFHNet({Q}_t)$  \Comment{Grasp Synthesis}
        \State $g^{*}_{t} \gets FilterWithMetrics(\mathbb{G}_i)$
        \State $\overline{\mathbf{v}}_{G,t} \gets VelocityEstimate({Q}_t$, ${Q}_{t-\tau})$ 
    \Else \Comment{No Control/Visual Feedback}
        \State update $g^*_t$, ${Q}_t$ based on $g^{*}_{t-1}$, $Q_{t-1}$, $\overline{\mathbf{v}}_{G,t-1}$
        \State $\overline{\mathbf{v}}_{G,t} \gets \overline{\mathbf{v}}_{G,t-1}$
    \EndIf
        \State $HandPose_{t} \gets RobotControl(g^*_t$, $\overline{\mathbf{v}}_{G,t}$)
        \State $Score \gets GraspEval(HandPose_{t}$, $g^{*}_{t}$, ${Q}_t)$
    \If{$Score > thresh$} \Comment{Grasp Execution}
        \State $CloseFinger(g^*_t)$
        \State $\bf{break}$
\EndIf
\EndWhile
\EndFunction
\end{algorithmic}
\end{algorithm}

If any of these conditions is violated, the recent observation is discarded. Otherwise, the model point cloud, as well as the previous $n_s$ observation point clouds $\mathbf{P}_{t-\tau}$, are transformed into the current camera frame to match the pose of the recent observation point cloud $\mathbf{P}_t$. Here $\tau = 1,2,..,n_s$. \par
As an additional post-processing step, the result is smoothed by selecting only points with correspondences in the previous $n_s$ observations. When discretizing the continuous surface of the target object with a point cloud, the locations of single points in consecutive time steps differ due to the discretization grid depending on the camera and object pose. Therefore, a corresponding point is defined as one that lies in an $\epsilon$-ball around the source point. Only if at least one point $ p'_{t-\tau,j}$ can be found in every previous $n_s$ observation point clouds $\mathbf{P}'_{t-\tau}$ with an Euclidean distance to the source point lower than a small positive constant $\epsilon$, this source point $p'_{t,i}$ is considered for further processing
\begin{equation}
\mathbf{P}_t'' = \bigcup\limits_{\mathbf{P}_t'} p'_{t,i} [\forall \tau: \exists p'_{t-\tau,j} \in \mathbf{P}'_{t-\tau}: ||p'_{t,i} - p'_{t-\tau,j}|| < \epsilon].
\end{equation}

Finally, the post-processed observation point cloud $\mathbf{P}_t''$ is merged with the model point cloud $\mathbf{Q}_{t-1}$ to obtain an improved model point cloud $\mathbf{Q}_{t}$
\begin{equation}
\mathbf{Q}_{t} = \mathbf{Q}_{t-1} \cup \mathbf{P}_t''.
\end{equation}

The model point cloud is regularly downsampled by removing randomly selected points. This keeps the number of points under control and reduces the amount of outliers over time, as it is unlikely that they will erroneously appear in the model point cloud again once they have been removed.\par
To allow efficient processing of the model point cloud for machine learning algorithms, it is encoded with the Basis Point Set (BPS)~\cite{Pro19}. The points are represented by their 1D distance to each fixed basis point instead of their 3D coordinates.

\subsection{Grasp Control}\label{gg}
The BPS-encoded model point cloud generated by the \textit{Target Model Generation} process is fed to \textit{Grasp Control} process, which constitutes the input for the generative grasping framework \textit{Five-finger Hand Net} (\textit{FFHNet})~\cite{May21}. It samples a set of grasps and assigns real-time success predictions $s_{G,i}$. 
Every grasp is represented by its palm translation $\mathbf{t}_{G,i}$, its palm rotation matrix $\mathbf{R}_{G,i}$ and its finger configuration $\bm{\theta}_{G,i}$.\par

The following paragraphs describe the post-processing steps for selecting a suitable grasp and controlling the robot's movement to reach a pose that is predicted as a successful grasp. Further steps are estimating the target's velocity and updating the grasp pose in case of missing control feedback due to tracking or ICP failure.

\subsubsection{Dynamic Grasping Metric}~\label{grasp_metric}
From the generated grasps described by the set $\mathbb{G}_i = (\mathbf{t}_{G,i}, \mathbf{R}_{G,i}, \bm{\theta}_{G,i}, s_{G,i})$, one grasp at time step $t$ is selected to maximize a specifically designed metric
\begin{equation}
g^*_t = \argmax_i{\sum_j m_j(\mathbb{G}_i)}.
\end{equation}
This metric combines the following quantities:\par
\textbf{Semantic Cue -- Predicted grasp success:} By semantic cue, we mean the knowledge about what is a good grasp learned by the grasp evaluator. 
The evaluator has been trained to implicitly incorporate more information than just the geometric one, such as feasibility and collision avoidance of the grasp.  
This part is weighted with the constant $c_{m,s}$
\begin{equation}
m_0(\mathbb{G}_i) = c_{m,s} s_{G,i}.
\end{equation}\par
\textbf{Geometric Cue -- Pose difference:} This part of the metric attempts to minimize a pure geometric quantity, that is the Euclidean distance between the robot's pose $(\mathbf{t}_{R}, \mathbf{r}_{R})$ and the grasp pose $(\mathbf{t}_{G,i}, \mathbf{r}_{G,i})$ in axis-angle representation. The linear and angular offset are weighted with the constants $c_{m,t}$ and $c_{m,r}$
\begin{equation}
m_1(\mathbb{G}_i) = - c_{m,t} || \mathbf{t}_{G,i} - \mathbf{t}_{R} || - c_{m,r} || \mathbf{r}_{G,i} - \mathbf{r}_{R} ||.
\end{equation}
Shorter robot movements reduce the grasp execution time. This increases efficiency and mitigates the failure risk caused by a target movement out of the robot's reachable space.\par

The grasps ranked by dynamic grasping metrics will be iteratively filtered with kinematic reachability until one reachable grasp is found.

\subsubsection{Robot Control}
After evaluation of the grasp metric, the robot is moved towards the pose $(\mathbf{t}_G^*, \mathbf{R}_G^*)$ of the selected grasp $g^*_t$. 
We choose the eye-in-hand camera setup to reduce the risk of hand-target collisions since the object is approached with an open palm. The remaining limitation is that the object must remain in the camera's field of view to provide visual feedback for the control loop. Therefore, the control law for the end-effector's orientation is adjusted to face the object as long as possible, before aligning the desired grasp orientation. 

The desired behavior is achieved by combining two control targets for the end-effector orientation depending on the translation error $||\mathbf{t}_G^{*}||$. 
As long as the translational error's norm of the end-effector position $||\mathbf{t}_G^{*}||$ is larger than a threshold $c_{O}$, the camera is aligned towards the target's center. 
Once this threshold is undershot, the desired orientation is linearly interpolated between the grasp orientation error $\mathbf{r}_G^*$ and the alignment error between the camera center and object center $\mathbf{r}_{O}$. When $||\mathbf{t}_G^{*}||$ falls below another threshold $c_{G}$, exclusively the target grasp's orientation $\mathbf{r}_G^*$ is approached.

\begin{align}
&\tilde{\mathbf{r}}_G^* = \left\{\begin{array}{cl} \mathbf{r}_{O}, & ||\mathbf{t}_G^{*}|| \geq c_{O}\\
\Delta g \cdot \mathbf{r}_{O} + (1 - \Delta g) \cdot \mathbf{r}_G^*, &
||\mathbf{t}_G^{*}|| \in
(c_{G}, c_{O})\\
\mathbf{r}_G^* & ||\mathbf{t}_G^{*}|| \leq c_{G}
\nonumber
\end{array}\right.\\
&\mbox{with } \Delta g = \frac{||\mathbf{t}_G^{*}|| - c_{G}}{c_{O} - c_{G}} \mbox{, } c_{O} > c_{G} \mbox{.}
\end{align}

All rotations are represented as axis-angle vectors.\par
With the resulting rotation $\tilde{\mathbf{r}}_G^*$, the desired Cartesian end-effector velocity $(\mathbf{v}_{d}, \bm{\omega}_{d})^\top$ is calculated. 
\begin{equation} \label{v_control}
\begin{pmatrix} 
\mathbf{v}_{d} \\
\bm{\omega}_{d}
\end{pmatrix} = 
\begin{pmatrix} 
c_{p,v} \mathbf{t}_G^* + c_{d,v} \dot{\mathbf{t}}_G^* + \overline{\mathbf{v}}_{G,t} \\
c_{p,\omega} \tilde{\mathbf{r}}_G^* + c_{d,\omega} \dot{\tilde{\mathbf{r}}}_G^*
\end{pmatrix}
\end{equation}
$\overline{\mathbf{v}}_{G,t}$ is the estimated velocity of the target at time step $t$. The PD controller with the empirically tuned control constants $c_{p,v}$, $c_{d,v}$, $c_{p,\omega}$, $c_{d,\omega}$ ensures fast grasp pose alignment. It also compensates for changes in translation or rotation of the target grasp if the object moves.  

\subsubsection{Velocity Estimation}~\label{vel_est}
The evolution of the target's position is tracked based on a fixed feature point on its surface. The object's velocity $\overline{\mathbf{v}}_{G,t}$ at time step $t$ is estimated by processing the position of this feature point over time with a Kalman filter~\cite{Kal60}.
The estimated velocity is included in the control law in Equation~\ref{v_control}. 
This allows the controller to behave similarly to a static grasping setup, assuming an accurate velocity estimate $\overline{\mathbf{v}}_{G,t}$.
Additionally, this velocity is required to estimate a suitable grasp pose without useful control feedback.\par
We have found that estimating the rotational velocity does not improve the grasp success rate for two reasons: Firstly, the noisy observations cannot always be aligned with high precision, so a reliable and accurate estimation of the rotational velocity is not achieved. Secondly, mapping object rotation to a rotation of the target grasp (as done for translation) often results in singular robot configurations or collisions with the environment. Therefore, we assume no significant rotation of the object during tracking loss.

\subsubsection{Grasp Pose Update}
When the control feedback is missing, the target grasp pose is updated by moving it according to the estimated object velocity. Here, we assume the object moves with the same velocity during this short period.
The control feedback can be missing for several reasons: the limited camera's field of view and depth range, as well as the wrong transformation from the ICP algorithm. Especially shortly before grasp execution, the risk of point cloud registration failure is increased when the object is close to the camera. The camera's minimum depth range can lead to incorrect depth values, and its limited field of view can result in incorrect ICP registration because the object is only partially visible. For example, a corner of a partially observed box can be aligned with any box corners since their structure is similar.\par

\subsubsection{Grasp Execution}
We close the fingers when the current robot pose is likely to result in a successful grasp. 
This is assumed to be more effective than closing the fingers when the robot reaches the predicted grasp pose.
Because there exist many equally successful grasps, we can detect them with the grasp evaluator network in our framework, which has already learned exactly how to detect the stable grasp we need. 
In detail, the grasp evaluator network outputs a grasp score for each current robot hand pose $HandPose_{t} = (\mathbf{t}_{R}, \mathbf{r}_{R})$, combined with the finger configuration from predicted grasp $g^*_t$ and the model point cloud ${Q}_t$. 
This score is compared to a threshold $c_e$ to determine the execution. 


\section{Experiments}

Two experiments were carried out to demonstrate the effectiveness of our framework. In the first, the target objects are grasped on a conveyor belt, representing an industrial setting: productivity increases if the conveyor belt does not need to halt during object picking. The linear velocities between \SI{0}{} and \SI{220}{mm \per s} of the conveyor belt, as well as the direction of movement, are unknown to the system and must be estimated online.\par
The second experiment is a human-to-robot handover, which is highly relevant to human-robot interaction. Naturally, humans act cooperatively during a handover by moving the object towards the robot and slowing down when the object is within the robot's reach. However, a human-to-robot handover cannot be achieved with static grasping since the target grasp pose is a priori unknown, and humans cannot hold the target object perfectly still.\par
\begin{figure}[t]
\vspace{5pt}
\includegraphics*[width=\columnwidth]{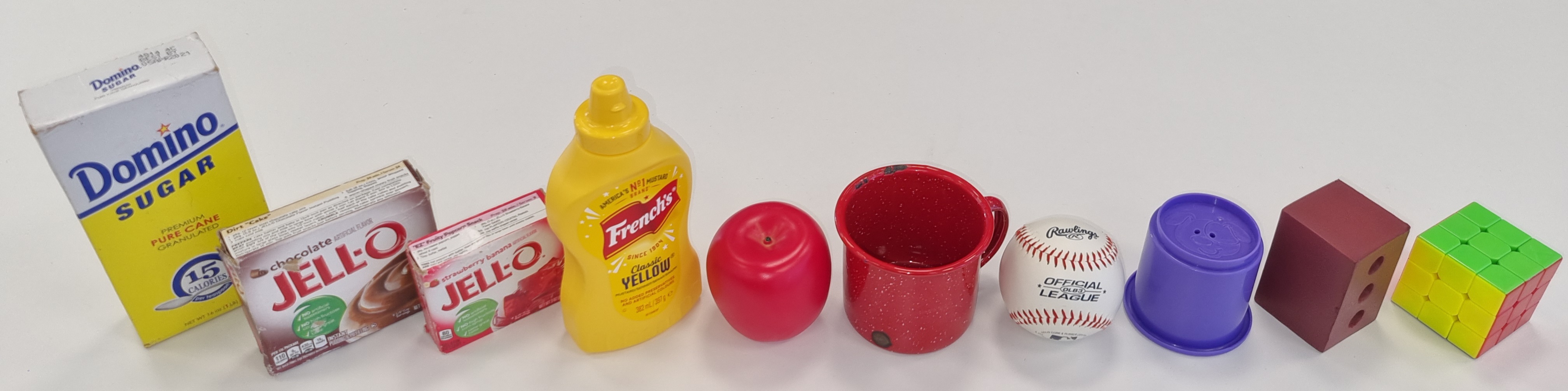}
\centering
\caption{The ten YCB objects used in the experiments (left to right): sugar box, pudding box, gelatin box, mustard bottle, apple, mug, baseball, cup, foam brick, Rubik's cube.}
\label{objects}
\end{figure}

Figure~\ref{exp_photos} and the \href{https://youtu.be/b87zGNoKELg}{accompanying video} show grasps from the experiments. The target items are displayed in Figure~\ref{objects}. They are a subset of the \textit{Yale-CMU-Berkeley} (\textit{YCB}) object set~\cite{ycb}, which was omitted during the training of our grasp generation module. Therefore, all of the target objects are unknown to the system.\par
The hardware used for the experiments is the 7 DoF robot manipulator \textit{Agile Robots Diana 7}, the \textit{DLR-HIT Hand II}~\cite{Liu08}, and the depth camera \textit{Intel RealSense D435}. These components are controlled by a computer equipped with a \textit{NVIDIA GeForce RTX 3070 Ti} graphics card.\par
The system parameters were empirically tuned before the experiments. In the Appendix, Table \ref{parameters} lists their values.

\subsection{Grasping Objects on a Conveyor Belt}
In this experiment, the ten target objects are grasped while moving linearly on a conveyor belt. Every object is attempted once for every speed increment of \SI{20}{mm \per s} between \SI{0}{} and \SI{220}{mm \per s}.

\textbf{Analysis of success rate over speed:}
Table~\ref{diagram} shows the grasping success rates for the respective conveyor belt velocities. For \SI{120}{} grasp attempts, an average success of \SI{71.7}{\%} is achieved. The diagram indicates that the system works reliably with success rates of \SI{80}{}--\SI{90}{\%} for target velocities up to \SI{20}{mm \per s} and between \SI{140}{} and \SI{200}{mm \per s}. Impressively, there is no performance degradation at these high speeds compared to static grasping. The success rate for \SI{220}{mm \per s} decreases because the limits of the system's capabilities are reached.\par

However, the success rates also drop to \SI{50}{}--\SI{60}{\%} for target speeds between \SI{60}{} and \SI{100}{mm \per s}. This effect can be explained by the construction of the target model. The model point cloud primarily comprises recent observations due to regular downsampling. When grasping a slowly moving object, all recent observations are captured from the same perspective, which resulFts in a less complete target model than for faster velocities. The grasp generation module was trained for static grasping from one perspective and worked reliably for (quasi) static objects. However, a more complete target model increases the quality of generated grasps. This increased reliability is required by the system to overcome the additional disturbances when grasping moving targets.\par


\begin{table}[t]
\begin{center}
\vspace{1pt}
\caption{Results of the conveyor belt experiment.\label{diagram}
}
\begin{tabular*}{\linewidth}{@{\extracolsep{\fill}}lcccccc}
\toprule
Speed [mm/s] & 0 & 20 & 40 & 60 & 80 & 100 \\
Success rate & \SI{90}{\%} & \SI{90}{\%} & \SI{70}{\%} & \SI{50}{\%} & \SI{60}{\%} & \SI{50}{\%}\\
\midrule
Speed [mm/s] & 120 & 140 & 160 & 180 & 200 & 220 \\
Success rate & \SI{70}{\%} & \SI{80}{\%} & \SI{80}{\%} & \SI{90}{\%} & \SI{80}{\%} & \SI{50}{\%} \\
\bottomrule
\vspace{-12pt}
\end{tabular*}
\end{center}
\end{table}

\textbf{Analysis of success rate per object:}
Table~\ref{tab_results} lists the success rates for each target object. High performance is achieved for many objects with varying shapes and sizes. Large cuboid objects (sugar box), sphere-like objects (apple) as well as smaller cuboid (foam brick) and cylindrical objects (cup) are grasped successfully in more than \SI{80}{\%} of the attempts. 
The objects with the worst performance are the mustard bottle, the mug, and the baseball. The shape of the mustard bottle causes the grasp generation module to often generate grasps for its small lid, requiring extremely high precision. Due to the mug's great diameter, any deviation from the grasp pose can lead to collisions. Finally, the spherical shape of the baseball does not provide any structural features to align different perspectives. That is why the resulting target model represents only a fraction of the real surface, resulting in unreliable grasp predictions.\par 

\textbf{Overall failure analysis:}
In Table~\ref{tab_fail}, the grasp failures are divided into the three cases \textit{imprecise grasp pose}, \textit{hand-target collision}, and \textit{bad grasp timing}. An imprecise grasp pose resulting in an unstable grasp or miss of the target accounts for almost every second grasp failure. This case occurred most often for the mustard bottle and the baseball because of an error-prone or inaccurate grasp generation for these object shapes. The reason for roughly a third of the failures is a collision of the hand with the target, most commonly observed for the mug. Its large diameter requires exact velocity estimation, grasp prediction, and robot control. The remaining failures are caused by a suitable hand pose not being recognized by the grasp evaluator and the grasp not being executed at the right time. The reason can be a false success prediction or inaccuracies in constructing the virtual target model. No object is particularly affected by this case.\par
Although we have seen some side grasps during system development, all grasps in the experiments were executed from the top. While our system is not limited to a specific grasp orientation, FFHNet has a strong top-grasp bias. In its training process, grasps colliding with a virtual bottom plane were classified as unsuccessful, resulting in higher success score predictions for top grasps.\par
\begin{figure}[t]
\includegraphics*[width=\columnwidth]{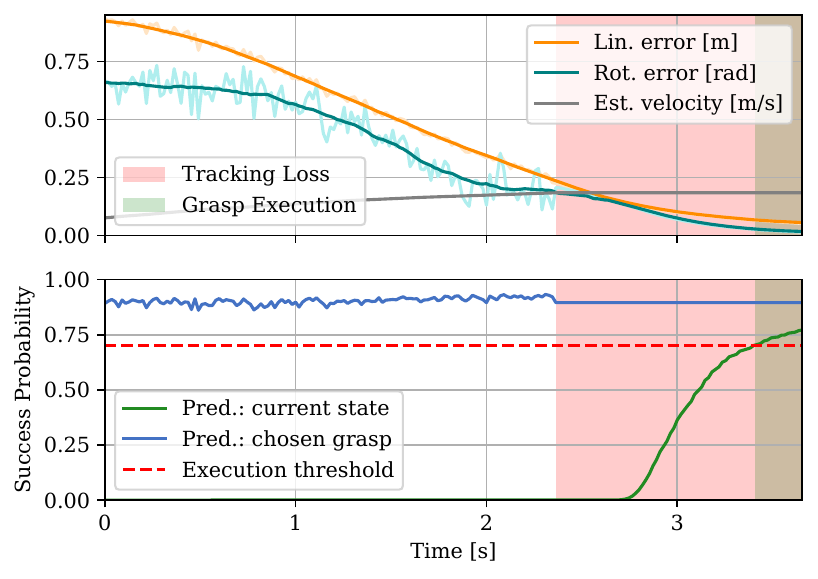}
\centering
\caption{Evolution of system quantities when successfully grasping the sugar box at \SI{200}{mm \per s} on the conveyor belt.}
\vspace{-2pt}
\label{dynamic_graph}
\end{figure}

\textbf{System quantity analysis:}
The plots in Figure~\ref{dynamic_graph} show some of the system's quantities to grasp the sugar box on the conveyor belt. The linear (lin.) and rotational (rot.) errors are smoothed for better visibility and strive towards zero. The estimated (est.) velocity requires some time to converge to the ground truth value of \SI{200}{mm \per s} since the conveyor belt is started after the system. After \SI{2.4}{s}, the visual feedback is missing, and the system relies on its estimation. Tracking loss commonly occurs for large objects like the sugar box in the final approach phase because the ICP algorithm fails to reliably align the partial camera observations with the constructed virtual model. As visible in the lower plot, the success prediction (pred.) for the currently estimated state converges to the success prediction of the chosen grasp until the execution threshold is crossed. Without visual feedback, the system's grasp pose estimation still allows a successful grasp execution.\par

\begin{table}[t]
\begin{center}
\vspace{5pt}
\caption{Success rates of the target objects.\label{tab_results}}
\begin{tabular*}{\linewidth}{@{\extracolsep{\fill}}lccc}
\toprule
 & Conveyor belt & Handover & Combined \\
\midrule
Sugar box & \SI{83.3}{\%} & \SI{80}{\%} & \SI{81.8}{\%} \\

Pudding box & \SI{75}{\%} & \SI{100}{\%} & \SI{86.4}{\%} \\

Gelatin box & \SI{66.7}{\%} & \SI{100}{\%} & \SI{81.8}{\%} \\

Mustard bottle & \SI{41.7}{\%} & \SI{50}{\%} & \SI{45.5}{\%} \\

Apple & \SI{100}{\%} & \SI{60}{\%} & \SI{81.8}{\%} \\

Mug & \SI{50}{\%} & \SI{90}{\%} & \SI{68.2}{\%} \\

Baseball & \SI{58.3}{\%} & \SI{70}{\%} & \SI{63.6}{\%} \\

Cup & \SI{83.3}{\%} & \SI{70}{\%} & \SI{77.3}{\%} \\

Foam brick & \SI{91.7}{\%} & \SI{60}{\%} & \SI{77.3}{\%} \\

Rubik's cube & \SI{66.7}{\%} & \SI{90}{\%} & \SI{77.3}{\%} \\

Overall & \SI{71.7}{\%} & \SI{77}{\%} & \SI{74.1}{\%} \\
\bottomrule
\vspace{-12pt}
\end{tabular*}
\end{center}
\end{table}

\begin{table}[t]
\begin{center}
\caption{Failure case analysis.\label{tab_fail}}
\begin{tabular*}{\linewidth}{@{\extracolsep{\fill}}lccc}
\toprule
 & Conveyor belt & Handover & Combined \\
\midrule
Imprecise grasp pose & \SI{47.1}{\%} & \SI{82.6}{\%} & \SI{61.4}{\%} \\

Hand-target collision & \SI{35.3}{\%} & \SI{17.4}{\%} & \SI{28.1}{\%} \\

Bad grasp timing & \SI{17.6}{\%} & \SI{0}{\%} & \SI{10.5}{\%} \\
\bottomrule
\vspace{-12pt}
\end{tabular*}
\end{center}
\end{table}

\subsection{Human-to-Robot Handover}
In this experiment, ten participants from our lab are asked to hand the target objects to the robot. 
In case of tracking loss, the system assumes a constant target pose ($\overline{\mathbf{v}}_{G} = \mathbf{0}$).\par

\textbf{Analysis of success rate per object:}
As listed in Table~\ref{tab_results}, the overall success rate is \SI{77}{\%} for \SI{100}{} grasp attempts. Compared to the conveyor belt experiment, the success rate is increased because humans present a handover object in a way that facilitates grasping for the receiver. Consequently, many objects are grasped at success rates of \SI{90}{\%} or higher: the pudding box, the gelatin box, and the Rubik's cube as small and midsize cuboid objects, but also the large cylindrical mug. Interestingly, the latter is the object with the second-worst result in the conveyor belt experiment. In contrast to the conveyor belt setting, the humans' cooperative behavior supports a precise enclosure of the object between the robotic fingers. Conversely, for the apple object and the foam brick, only \SI{60}{\%} of the handovers are successful, with them being the objects with the highest success rates in the conveyor belt experiment. The executed grasps for these items seem to be confusing for the human participants. In both experiments, the system performs worst on the mustard bottle.\par

\textbf{Overall failure analysis:}
Table~\ref{tab_fail} indicates an imprecise grasp pose as the central failure reason. Similar to the conveyor belt experiment, this occurs most commonly for the mustard bottle because of an error-prone grasp generation for the unconventional shape. The small number of hand-target collisions can be explained by humans trying to avoid them during a handover. Most of these failures occurred when handing over the apple, as some of the system's grasping attempts may have been misleading to the human participants. A delayed execution never led to grasp failure in this experiment.\par
Figure~\ref{handover_graph} shows the minimization of the linear and rotational errors and the increase of the success prediction for a successful handover of the Rubik's cube without significant tracking loss.\par

\afterpage{
\begin{figure}[t]
\includegraphics*[width=\columnwidth]{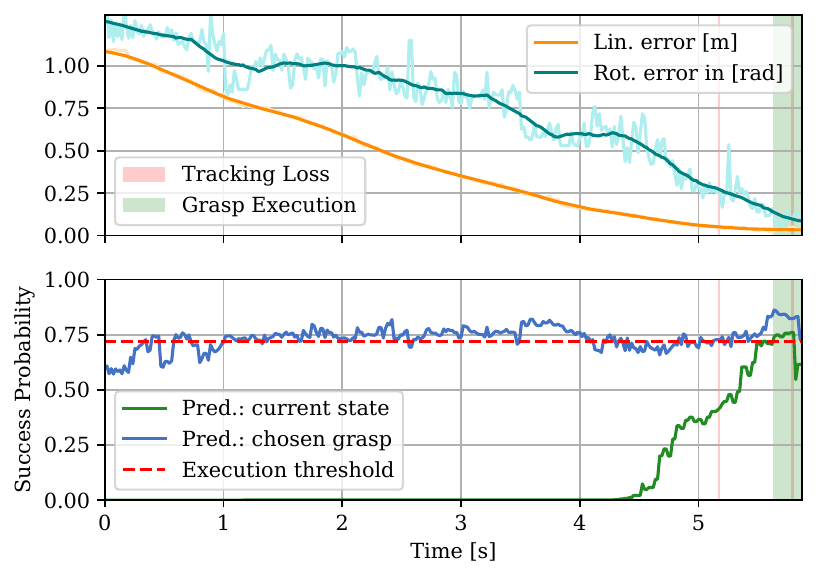}
\centering
\caption{Evolution of the system quantities for a successful handover of the Rubik's cube.}
\vspace{-2pt}
\label{handover_graph}
\end{figure}
}
\subsection{Limitations}
The reasons for the most common failure case, a grasp execution with an imprecise hand pose, are insufficient target model creation and inaccurate grasp predictions. To further improve the system's precision, a more robust approach to construct the target model than visual object tracking + ICP and a more reliable grasp generation is required. \par
Additionally, generating a less restrictive grasp data set and retraining FFHNet will overcome its top-grasp bias. With a model capable of generating a diverse distribution of suitable grasps, the grasp metric can also be extended to evaluate the direction of the target movement. Approaching the target from the direction in which it is moving can increase the success rate since it facilitates dynamic grasping~\cite{Aki21}.\par
The remaining failure cases due to hand-target collisions can be addressed by implementing capable trajectory planning and collision checking in real time. However, this requires a reliable creation of the virtual target model.

\section{Conclusion}
We presented a system to grasp unknown dynamic objects with a multi-fingered hand.
It constructs a virtual target model by processing camera data from a single view with a visual object tracking model, ICP, and further filtering. 
This model constitutes the input to the generative grasping model FFHNet, providing a distribution of possible grasps. Then the most suitable grasp is selected and continuously updated in real time. 
The virtual model and velocity estimation enable the system to compensate for missing visual feedback in case of tracking loss or alignment errors of the observed camera data. 
In our conducted experiments with real hardware, the system grasped various dynamic objects in two challenging scenarios with a high success rate.
As there is no prior work that can build such a framework for this complex task, we hope that our foremost investigation can pave the way to a robust and performant robotic system that can be deployed in the actual production line in the future.
\par

\bibliographystyle{IEEEtran}
\bibliography{IEEEabrv, bibliography}

\appendix

\section{Parameter Values}
\begin{table}[ht]
\begin{center}
\caption{Values of system parameters\label{parameters}}
\begin{tabularx}{\columnwidth}{l@{} S@{\hspace{10pt}} X}
\toprule
Parameter & {Value} & Description \\
\midrule
$c_z$ & 0.1 \SI{}{m} & Maximal deviation from median depth \\
$c_{a,f}$ & 0.8 & Minimal ICP alignment fitness score \\
$c_{a,\hat{v}}$ & 4.0\SI{}{m \per s} & Linear velocity limit between camera and target for valid alignment \\
$c_{a,\hat{\omega}}$ & 4.0\SI{}{rad \per s} & Rotational velocity limit between camera and target for valid alignment \\
$n_s$ & 5 & Number of previous observations considered for filtering \\
$\epsilon$ & 0.01\SI{}{m} & Correspondence radius for two points \\
$c_{m,s}$ & 1.0 & Metric weight for success score \\
$c_{m,t}$ & -0.1 & Metric weight for translational distance \\
$c_{m,r}$ & -0.2 & Metric weight for rotational distance \\
$c_O$ & 0.2\SI{}{m} & Distance threshold for camera orientation towards object \\
$c_G$ & 0.05\SI{}{m} & Distance threshold for grasp pose alignment \\
$c_{p,v}$ & 1.0 & Proportional control constant (translation) \\
$c_{d,v}$ & 2.0 & Derivative control constant (translation) \\
$c_{p,\omega}$ & 2.5 & Proportional control constant (rotation) \\
$c_{d,\omega}$ & 5.0 & Derivative control constant (rotation) \\
$c_e$ & 0.7 & Grasp execution threshold \\
\bottomrule
\end{tabularx}
\end{center}
\end{table}
\end{document}